\documentclass[10pt,conference]{IEEEtran} 
\IEEEoverridecommandlockouts
\usepackage{cite}
\usepackage{amsmath,amssymb,amsfonts}
\usepackage{algorithmic}
\usepackage{graphicx}
\usepackage{textcomp}
\usepackage{xcolor}
\usepackage{hyperref}
\usepackage{tabularx}
\usepackage[most]{tcolorbox}
\usepackage{booktabs}
\usepackage{makecell}
\usepackage{pgfplots}
\pgfplotsset{compat=1.16}
\usepgfplotslibrary{polar}
\usepackage{subcaption}
\definecolor{amagent}{HTML}{1A80BB}
\definecolor{codexcol}{HTML}{7F77DD}
\definecolor{ccodecol}{HTML}{888780}
\definecolor{gencol}{HTML}{D8A6A6}
\definecolor{noevocol}{HTML}{BA7517}
\usepackage{pifont}
\newcommand{\cmark}{\ding{51}}
\newcommand{\xmark}{\ding{55}}
\makeatletter
\newcommand{\linebreakand}{%
  \end{@IEEEauthorhalign}
  \hfill\mbox{}\par
  \mbox{}\hfill\begin{@IEEEauthorhalign}
}
\makeatother

\def\BibTeX{{\rm B\kern-.05em{\sc i\kern-.025em b}\kern-.08em
    T\kern-.1667em\lower.7ex\hbox{E}\kern-.125emX}}

\begin{document}

\title{Validation-Gated ML Pipeline Synthesis with Expert-Grounded Self-Evolving Skills}

\author{
% ==================== Row 1 ====================

\IEEEauthorblockN{Zixiao Zhao}
\IEEEauthorblockA{
\textit{School of Computer Science}\\
\textit{University of Auckland}\\
Auckland, New Zealand\\
zixiao.zhao@auckland.ac.nz
}
\and
\IEEEauthorblockN{Jing Sun}
\IEEEauthorblockA{
\textit{School of Computer Science}\\
\textit{University of Auckland}\\
Auckland, New Zealand\\
jing.sun@auckland.ac.nz
}
\and
\IEEEauthorblockN{Zhe Hou}
\IEEEauthorblockA{
\textit{School of Information and}\\
\textit{Communication Technology}\\
\textit{Griffith University}\\
Brisbane, Australia\\
z.hou@griffith.edu.au
}
\and
\IEEEauthorblockN{Mengze Li}
\IEEEauthorblockA{
\textit{School of Computer Science}\\
\textit{University of Auckland}\\
Auckland, New Zealand\\
mengze.li@auckland.ac.nz
}

\linebreakand

% ==================== Row 2 ====================

\IEEEauthorblockN{Qian Liu}
\IEEEauthorblockA{
\textit{School of Computer Science}\\
\textit{University of Auckland}\\
Auckland, New Zealand\\
liu.qian@auckland.ac.nz
}
\and
\IEEEauthorblockN{Cheng-Hao Cai}
\IEEEauthorblockA{
\textit{Metasics, China.}\\
Xiamen, China\\
chenghao.cai@metasics.cn
}
\and
\IEEEauthorblockN{Zijian Zhang}
\IEEEauthorblockA{
\textit{School of Cyberspace}\\
\textit{Science and Technology}\\
\textit{Beijing Institute of Technology}\\
Beijing, China\\
zhangzijian@bit.edu.cn
}

% ==================== Row 3 ====================
\linebreakand

\IEEEauthorblockN{Xin Wang}
\IEEEauthorblockA{
\textit{DSA Thrust}\\
\textit{The Hong Kong University of}\\
\textit{Science and Technology}\\
\textit{(Guangzhou)}\\
Guangzhou, China\\
xwang496@connect.hkust-gz.edu.cn
}
\and
\IEEEauthorblockN{Gillian Dobbie}
\IEEEauthorblockA{
\textit{School of Computer Science}\\
\textit{University of Auckland}\\
Auckland, New Zealand\\
g.dobbie@auckland.ac.nz
}
\and
\IEEEauthorblockN{Jin Song Dong}
\IEEEauthorblockA{
\textit{Department of Computer Science}\\
\textit{National University of Singapore}\\
Singapore\\
dcsdjs@nus.edu.sg
}
}

\maketitle

\begin{abstract}
Recent advances in large language models have enabled agentic systems that automate machine learning pipeline construction from minimal user input. However, existing agentic AutoML methods either depend on rigid templates that limit adaptability, target narrow task domains, or rely on free-form code generation that lacks reliability guarantees, leaving intermediate errors to propagate silently and degrade the final model. We present EvoML, an agentic framework for end-to-end ML pipeline synthesis built on expert-grounded skills that evolve across runs, executed through a validation-gated harness. The framework decomposes pipeline construction into four phases coordinated by three agents: an Orchestrator that enforces validation gates, advancing each phase only when its outputs pass structural, execution, and semantic checks, which precludes error propagation and keeps every artifact auditable; an Executor that runs expert-grounded skills, grounded in the nnU-Net workflow and pairing reusable templates with online adaptation; and a Curator that performs post-run evolution, consolidating the scripts and guidance recovered during a run into the skill library so it improves across datasets. We evaluate EvoML on eight datasets spanning six sophisticated, real-world ML task types, against agentic AutoML systems and command-line coding tools. EvoML achieves full automation and the best final model quality on all eight, spanning segmentation (up to 0.94 Dice), classification (0.99 accuracy), reconstruction (0.95 SSIM), and denoising (0.79 normalized PSNR), while transferring to broader ML tasks. 
\end{abstract}


\section{Introduction}
\label{sec:intro}
Automated Machine Learning (AutoML) aims to reduce the manual effort of building ML solutions by automating stages such as data preparation, model selection, hyperparameter optimisation, and pipeline construction~\cite{yao2018taking,he2021automl}. Traditional systems search over predefined model families, operators, and hyperparameter spaces~\cite{feurer2015efficient,olson2016tpot,drori2021alphad3m}, but their fixed search spaces and hand-designed primitives limit flexibility for complex workflows, domain-specific processing, or end-to-end executable code~\cite{barbudo2023eight,alsharef2022review,bahri2022automl}. Large language models (LLMs) and agentic coding systems offer a new opportunity to interpret high-level intent, generate executable code, invoke tools, and iteratively refine solutions~\cite{gu2025large}. Building on these capabilities, recent work applies LLMs to individual ML pipeline stages, including data and feature extraction~\cite{hong2025data,li2025exploring,malberg2024felix}, model selection and training~\cite{shen2023hugginggpt}, \cite{li2023traineragent}, and hyperparameter optimisation~\cite{liu2024large}, \cite{zhang2023using}, offering more flexible, context-aware automation than fixed search spaces.

This potential has further motivated agentic AutoML systems that aim to coordinate multiple ML stages within a single end-to-end workflow.
Recent systems~\cite{trirat2024automl,fang2026mlzero,feng2025m} demonstrate the potential of LLM-based agents for automating ML workflows, but several limitations remain when applied to robust end-to-end pipeline construction. 
First, they often require users to supply template code, dataset-specific features, or detailed solution descriptions beyond minimal problem input, making automation sensitive to the quality of human guidance. 
Second, execution is not consistently governed by effective validation gates that verify whether each stage has produced executable, reusable artifacts before subsequent stages proceed; as a result, failures can propagate through the workflow, leading to incomplete artifacts or requiring manual intervention to diagnose and correct. 
Third, pipeline construction is commonly driven by search over predefined operators, retrieved knowledge, or online examples, which can limit robustness when tasks demand domain-specific procedures or tight coordination across dependent stages. 
Finally, failures observed during execution are rarely converted into persistent improvements of the underlying procedures, leaving the system unable to accumulate operational experience across runs.

To address these limitations, we propose EvoML, a validation-gated ML pipeline synthesis framework with expert-grounded, self-evolving skills. 
EvoML decomposes end-to-end pipeline synthesis into three coordinated GPT-5.5~\cite{gpt5.5:online} agents. 
The Orchestrator governs stage-by-stage progression through the ML workflow, enforcing validation-gated transitions, advancing only when expected artifacts and validation signals satisfy predefined criteria, such as successful execution, required file generation, valid data format, or task-relevant metric constraints, and triggering retries.
The Executor performs each stage by invoking reusable, expert-grounded skills, each of which encodes a standard operating procedure for a specific ML stage, including data preparation, model configuration, training, hyperparameter optimisation, and code integration. 
The Curator analyses validation and retry traces after each run, identifies skills that incur repeated failures or excessive retries, and evolves those skills for future executions, allowing the framework to continuously refine its execution capabilities.
By separating workflow control, skill-based execution, and post-run skill evolution into distinct agent responsibilities, the framework supports autonomous pipeline synthesis from minimal user input while maintaining auditable validation, bounded failure recovery, and continuous improvement.

This paper makes the following contributions. 
\begin{itemize}
    \item We present \textbf{a fully automated agentic system for ML pipeline synthesis} that autonomously executes, monitors, and refines pipelines from minimal user input, without relying on task-specific templates or human guidance. 
    \item We introduce \textbf{a unified validation-gated execution harness} in which stage progression is governed by artifact- and metric-based checks, enabling robust transitions, bounded failure recovery, and auditable execution traces. 
    \item We couple \textbf{expert-grounded skills} with \textbf{post-run evolution} to support reliable and continuously improving ML pipeline synthesis, consolidating successful recoveries from validation and retry traces into \textbf{persistent skill updates across runs}.
\end{itemize}

The remainder of this paper is organised as follows: Section~\ref{sec:re_work} reviews previous methods on LLM-based program synthesis and agentic AutoML. 
Section~\ref{sec:methodology} presents the EvoML framework, detailing its three-agent design, expert-grounded skill library, validation-gated execution, and post-run skill evolution. 
Section~\ref{sec:evaluation} reports the experimental evaluation across a range of ML tasks, including ablation studies of the core components.
Section~\ref{sec:discussion} provides an architectural comparison with the baselines and discusses cost-efficiency and threats to validity. Section~\ref{sec:conclusion} concludes with a summary and future research directions.

% -------------------------------------------------------------
\section{Background}
\label{sec:re_work}
\textbf{LLM-Based Program Synthesis Frameworks.} Program synthesis, the automatic generation of programs from high-level specifications, is a long-standing software-engineering task~\cite{manna1971toward}; early formal, deductive approaches were limited in scalability and generality~\cite{gulwani2017program}. Large language models trained on code have since enabled the generation of syntactically correct, functionally plausible code from natural language~\cite{austin2021program,nijkamp2022codegen,jain2022jigsaw,jiang2026survey}, and agentic coding frameworks extend single-shot generation toward iterative, tool-augmented workflows~\cite{wang2025ai}, \cite{sapkota2025vibe}. Multi-agent systems such as MetaGPT~\cite{hong2024metagpt}, ChatDev~\cite{qian2024chatdev}, SWE-agent~\cite{yang2024swe}, and MaCTG~\cite{zhao2024mactg} decompose complex tasks across collaborating agents, and assistants such as GitHub Copilot~\cite{GitHubCo91:online}, Cursor~\cite{CursorTh1:online}, Codex~\cite{Introduc82:online}, and Claude Code~\cite{ClaudeCo70:online} bring these capabilities into practical use. These general-purpose frameworks, however, are not designed for the structural and procedural demands of ML pipeline construction: they lack domain-specific workflow coordination and depend on human-supplied guidance, such as task descriptions, template code, or annotated examples, limiting their applicability to autonomous end-to-end synthesis.

\textbf{Agentic AutoML Methods.} Applying agentic coding to ML pipeline synthesis raises challenges of stage coordination, domain-specific procedures, and execution reliability. Early efforts addressed individual stages in isolation: CAAFE~\cite{hollmann2023large} automates feature engineering and DS-Agent~\cite{guo2024ds} applies case-based reasoning to data processing and training, but both rely on structured human input and leave cross-stage coordination unaddressed. AutoML-GPT~\cite{zhang2023automl} and SELA~\cite{chi2024sela} attempt end-to-end automation through LLM-generated pipelines and tree search, yet still depend on detailed human input and lack validation-gated progression or failure recovery. The most capable end-to-end systems are AutoML-Agent~\cite{trirat2024automl}, which coordinates multiple agents but requires user-supplied task templates; M3Builder~\cite{feng2025m}, which performs strongly on medical imaging but is constrained to specific dataset types and tasks; and MLZero~\cite{fang2026mlzero}, which runs without manual input via documentation retrieval and iterative debugging but validates only execution success and retains memory within a single task. Across all of them, a common gap persists: none converts execution-time failures into persistent improvements, leaving the system unable to accumulate experience across runs.
% -------------------------------------------------------------
\section{Methodology}
\label{sec:methodology}

EvoML synthesises a complete, executable ML pipeline from a dataset and a minimal task description, without task-specific templates or human guidance. 
We follow a design-and-evaluate process in which the limitations identified in Sections \ref{sec:intro}–\ref{sec:re_work} are translated into three design requirements for reliable agentic ML pipeline synthesis: preventing error propagation across stages, grounding execution in reliable domain procedures, and retaining successful recovery experience across runs. 
These requirements motivate EvoML’s three core mechanisms: validation-gated orchestration, expert-grounded skill execution, and post-run skill evolution.

Accordingly, Section \ref{sec:methodology} first presents the overall system architecture and its three constituent agents, then introduces the expert-grounded skill library, describes the validation-gated execution that governs phase progression, and finally presents the post-run evolution mechanism that refines the skill library across runs.

\subsection{System Overview}
\label{sec:overview}

\begin{figure*}[!htbp]
    \centering
    \includegraphics[width=0.9\textwidth]{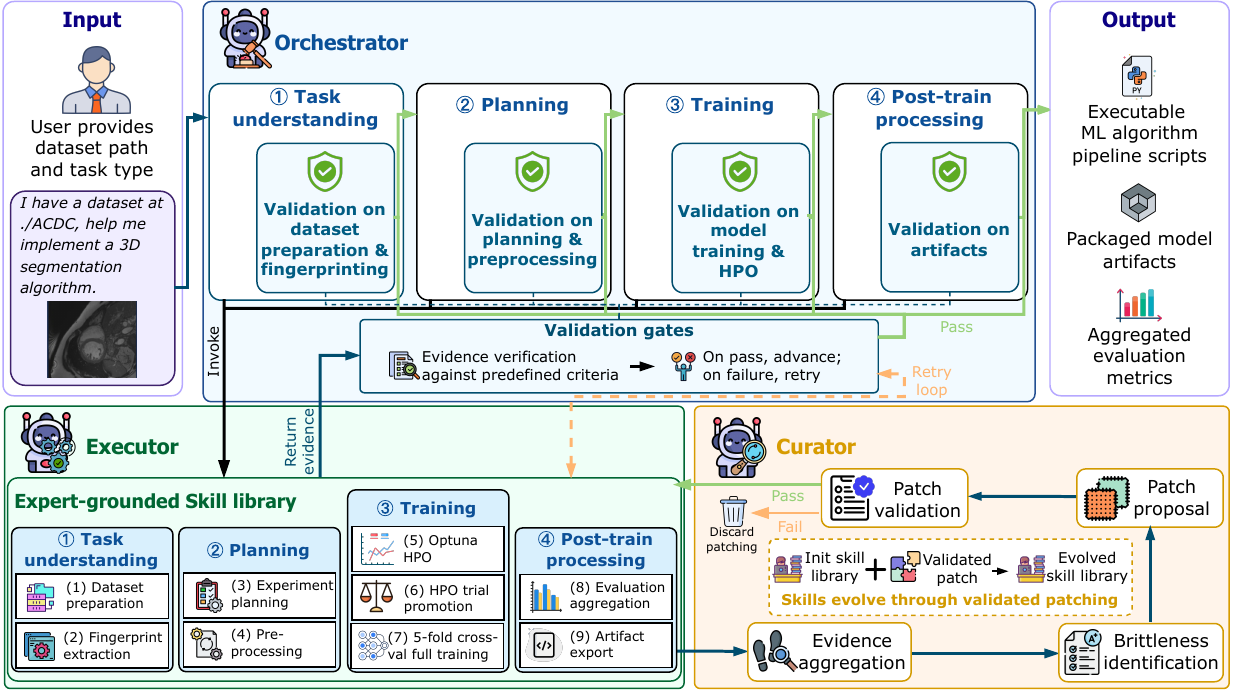}
    \caption{Overview of the proposed validation-gated agentic ML pipeline synthesis framework. The Orchestrator progresses through four phases and invokes the corresponding Executor skill group in sequence: Task Understanding (skills 1–2), Planning (3–4), Training (5–7), and Post-train Processing (8–9). After each phase, generated artifacts and execution evidence are returned for gated validation; passed phases advance, while failed phases trigger recovery. The Curator operates after the run to evolve the skill library from accumulated execution evidence.}
    \label{fig:pipeline}
    %\vspace{-8pt}
\end{figure*}

Figure~\ref{fig:pipeline} illustrates how the proposed framework addresses the ML pipeline synthesis task through three dedicated agents:
(1) the \textbf{Orchestrator}, which governs synthesis progression through validation gates;
(2) the \textbf{Executor}, which runs expert-grounded skill groups from the skill library for each phase; and
(3) the \textbf{Curator}, which operates post-run, auditing and evolving the skill library.
For an ML pipeline synthesis task, the input consists of a dataset and a minimal task description specifying the task type and data path. Given this, the system autonomously generates a complete, executable ML pipeline covering all stages from data preparation to model training and evaluation. EvoML organises the synthesis via these three coordinated agents across four main phases.

At runtime, the Orchestrator interprets the user request and drives the four phases in turn, invoking the Executor with the corresponding skill group at each: Task Understanding (dataset preparation and fingerprint extraction), Planning (experiment planning and preprocessing), Training (HPO based on Optuna~\cite{akiba2019optuna}, promotion, and full training), and Post-Train Processing (evaluation aggregation and artifact export). After each phase the Executor returns its artifacts and logs for gate evaluation; on failure the Orchestrator returns the failure context for a bounded retry, and on success it advances. Once the run concludes, the Curator inspects the accumulated evidence, identifies underperforming skills, and proposes verified updates to the library for future runs.

This separation of concerns gives each agent an isolated context and clear accountability at each phase boundary: the Orchestrator validates independently of the code that produced the artifacts, the Executor focuses solely on running skills, and the Curator analyses the complete execution evidence post-run rather than reacting with partial context mid-run. The four-phase structure mirrors the natural decomposition of an ML pipeline, with validation boundaries that localise failures before downstream computation is committed.

\subsection{Expert-Grounded Skill Library Design}
\label{sec:skills}
\subsubsection{Skill Structure and Expert Grounding}
Following the Agent Skills format~\cite{skills}, each skill in EvoML is a self-contained module comprising a Markdown specification, a set of Python script templates, and a reference document that encodes the relevant domain knowledge. 
The Markdown file encodes the stage objective, expected inputs, the usage of the accompanying Python templates, and a self-validation instruction for the Executor to verify successful execution. 
The reference document, predominantly an output-format specification such as the expected dataset format, grounds each skill's execution in a verifiable contract. 
Together, all skills form a shared library from which the Executor is prompted to invoke the appropriate skill at each stage of the current workflow.

Reliable end-to-end synthesis requires more than free-form code generation; it demands a principled procedure. We therefore ground EvoML's skills in nnU-Net~\cite{isensee2021nnu}, a self-configuring framework that achieves strong performance not through elaborate architectures but by analysing the dataset and adapting preprocessing, model, and training configuration to its characteristics. This data-driven, configuration-first approach is not specific to any one task type, which makes it a sound basis for general pipeline synthesis. This grounding also resolves a fundamental tension in agentic synthesis: pure LLM generation is flexible but structurally variable and unreliable across runs, whereas fixed templates are consistent but cannot capture task-specific requirements. Deriving templates from nnU-Net's well-defined procedures supplies structural consistency, while LLM-generated adaptations retain the flexibility each task demands. nnU-Net itself, however, requires manual deployment and expert configuration per task and lacks automated validation; EvoML resolves these limitations by encoding its workflow as agent-executable skills, automating configuration while adding validation-gated execution and cross-run evolution that nnU-Net lacks.

\subsubsection{Skill Library Composition}
The library comprises nine skills spanning the four phases. The \textit{Dataset preparation} skill converts a raw dataset into the standardised nnU-Net format, automatically selecting, cloning, and adapting the appropriate converter for the input layout. The \textit{Fingerprint extraction} skill then computes a compact statistical summary of the dataset, including image shape, voxel spacing, and intensity distribution, which conditions all subsequent configuration decisions. From this fingerprint, the \textit{Experiment planning} skill derives the model and training configuration, including network topology, patch size, and batch size, while the \textit{Preprocessing} skill crops, resamples to the target spacing, and normalises intensities to produce training-ready arrays.
The \textit{Optuna HPO} and \textit{HPO promotion} skills are introduced beyond nnU-Net's native workflow to improve final performance. The HPO skill runs hyperparameter optimisation trials on a fixed tuning fold, sampling learning rate, weight decay, and early-stopping configurations, while the promotion skill compares the best trial against the baseline configuration and promotes it only when the performance gain exceeds a predefined threshold. The \textit{Full training} skill then executes the standard nnU-Net five-fold cross-validation under the promoted configuration, augmented with an early-stopping mechanism that halts training once validation performance converges, and the \textit{Evaluation aggregation} skill consolidates the cross-validation predictions, determines postprocessing, and computes the final metrics. Finally, the \textit{Artifact export} skill packages the executed stage scripts into a standalone, runnable pipeline.
These nine skills constitute a fixed, complete decomposition of the ML pipeline; evolution therefore refines existing skills rather than inserting or deleting them.

\subsection{Validation-Gated Pipeline Execution}
\label{sec:execution}
The Orchestrator and Executor interact in a structured invoke-validate-advance loop within a unified validation-gated harness that enforces phase-level contracts, manages artifact dependencies, and provides bounded failure recovery across four phases from \textit{Task Understanding} to \textit{Post-Train Processing}.

\subsubsection{Phase Execution}
At each phase $p_k$, the Orchestrator invokes the Executor to run the skill group $\Phi_k$, which takes the artifact set $\mathcal{A}_{k-1}$ produced by the preceding phase as input and produces a new artifact set $\mathcal{A}_k$ according to:
\begin{equation}
\mathcal{A}_k = \Phi_k(\mathcal{A}_{k-1}, \mathcal{C}_k)
\end{equation}
where $\mathcal{C}_k$ denotes the task context at phase $p_k$, including the task type, run state, and any phase-specific configuration inherited from earlier phases.
Concretely, running $\Phi_k$ executes each constituent skill sequentially.

For each skill $\phi \in \Phi_k$, the Executor selects the appropriate Python template $\tau_k \in \phi$ based on the task context $\mathcal{C}_k$.
Where the template covers the required configuration, it is executed directly with task-specific parameters substituted; where adaptation is required, the Executor invokes Codex~\cite{Introduc82:online} as an MCP tool~\cite{Introduc8:online}, passing the template and task context to generate a task-specific adaptation. 
The Executor then executes the adapted script and verifies it against the self-validation criteria $v_k$ specified in the skill's Markdown, catching any runtime errors or tracebacks and re-invoking Codex with the error context to iteratively refine the adaptation until the script executes successfully. 
The Executor returns $\mathcal{A}_k$ to the Orchestrator only when $v_k(\mathcal{A}_k) = 1$, establishing self-validation as a necessary correctness condition that prevents structurally invalid outputs from reaching the phase-level validation gate and incurring unnecessary recovery overhead.

\subsubsection{Validation Gates and Recovery}
\label{sec:gate} 
Each gate verifies that the phase has produced the artifacts the next phase depends on: a formatted dataset and fingerprint in Task Understanding, an experiment plan and preprocessed data in Planning, the optimised configuration and trained weights in Training, and the aggregated metrics and exported pipeline in Post-Train Processing. At every phase, the Orchestrator applies the same three categories of check: structural (expected artifacts and formats), execution (process completion and exit status), and semantic (task-appropriateness of the artifacts). What differs between phases is not the checks but the reference they are evaluated against, which specifies the outputs a given phase should produce and remains largely stable across runs.

Structural and execution checks are performed programmatically, matching the produced artifacts against concrete expectations: the files that must exist, their formats and attributes, and each step's exit conditions. The semantic category admits no fixed rule: whether an artifact suits the task is a matter of convention, so the Orchestrator, as an LLM, evaluates it against the domain knowledge encoded in the phase's reference, for example applying clipping to CT rather than the z-score normalisation appropriate to MRI.

Formally, the gate evaluates the phase output $\mathcal{A}_k$ against the phase's criteria $\Gamma_k = \Gamma_k^{\mathrm{str}} \cup \Gamma_k^{\mathrm{exe}} \cup \Gamma_k^{\mathrm{sem}}$, collecting the structural, execution, and semantic criteria, and passes only if all are satisfied,
\begin{equation}
g_k(\mathcal{A}_k) = \bigwedge_{\gamma \in \Gamma_k} \mathbb{1}_\gamma(\mathcal{A}_k) \;\rightarrow\; \{0,1\} \times \mathcal{L}_k,
\end{equation}
where $\mathbb{1}_\gamma(\mathcal{A}_k) = 1$ if $\mathcal{A}_k$ satisfies criterion $\gamma$ and $0$ otherwise. For each failed criterion, the Orchestrator records its category $c_i$ and a natural-language summary $s_i$, producing the structured log
\begin{equation}
\mathcal{L}_k = \{(c_i, s_i) \mid \mathbb{1}_{\gamma_i}(\mathcal{A}_k) = 0\},
\end{equation}
returned to the Executor as $\mathcal{F}_k = (0, \mathcal{L}_k)$ for targeted recovery rather than full re-execution. A phase advances only when its gate passes; it retries while $r_k < R_{\max}$. If the retry limit is reached, the Orchestrator invokes a minimal-safe fallback that performs only the essential operations required by the phase contract without further optimisation. The fallback is evaluated by the same gate, and the workflow terminates if it also fails. In our experiments, normal recovery always succeeded before this fallback was required.
Because each phase executes only on artifacts that have passed all upstream gates, error propagation is structurally precluded, and every committed artifact is auditable through the criteria it satisfied.

\begin{figure}[!htbp]
    \centering
    \includegraphics[width=0.9\linewidth]{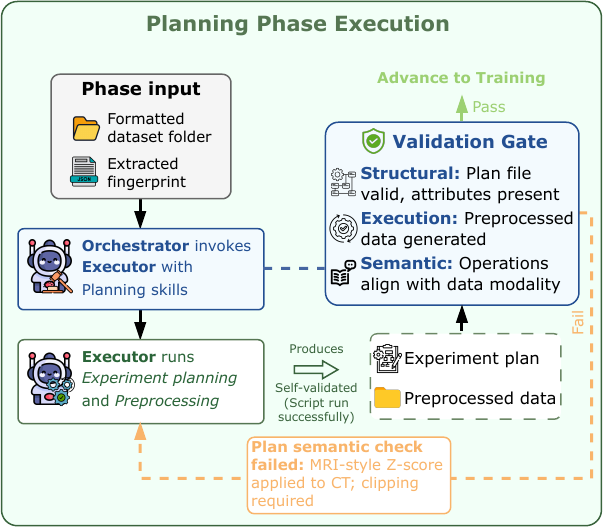}
    \caption{An example validation-gated recovery cycle in the Planning phase.}
    \label{fig:gate}
%\vspace{-8pt}
\end{figure}

Figure~\ref{fig:gate} illustrates an example cycle of this gating-recovery mechanism in the Planning phase. 
Taking the formatted dataset and the extracted fingerprint $\mathcal{A}_1$ as input, the Orchestrator invokes the Executor with the Planning skills. 
The Executor runs the experiment planning and preprocessing skills and self-validates that the scripts execute successfully, producing the experiment plan and preprocessed data as $\mathcal{A}_2$. 
The Orchestrator then evaluates $\mathcal{A}_2$ at the validation gate: although the structural and execution checks pass, the semantic check detects that a CT dataset has been normalised with an MRI-style Z-score rather than the required clipping, a correctness error that script-level self-validation cannot catch. 
The resulting failure context is returned to the Executor for a targeted retry, and only once all checks pass does the phase advance to Training.

\subsection{Post-Run Skill Evolution}
\label{sec:evo}
While the Orchestrator's recovery resolves failures within a run, its effects are yet transient: the adapted scripts it yields are not retained, and the guidance that misjudged the case remains unrefined. A skill encountering a similar condition must therefore rediscover the same resolution through retries, a pattern that recurs whenever such conditions arise. The Curator addresses this by operating post-run, transforming accumulated execution evidence into durable refinements of the skill library. In-run recovery and post-run evolution are therefore driven by a common signal: the validation and retry traces emitted at each gate. The harness persists these traces during execution, so the Curator's evolution draws on the same auditable evidence that governs in-run control.

\subsubsection{Evidence Aggregation and Brittleness Identification}
After a run concludes, the Curator aggregates the traces that the harness persisted for each phase into a per-skill evidence record. For a skill $\phi$, this record collects the retry count it incurred, the validation logs of the checks it failed, and the adapted script $\tau_\phi^\star$ that the Executor produced to resolve the difficult case during recovery:
\begin{equation}
\mathcal{E}_\phi = \left(r_\phi,\ \{\mathcal{L}_k \mid \phi \in \Phi_k\},\ \tau_\phi^\star\right), \qquad r_\phi = \sum_{k:\,\phi\in\Phi_k} r_k.
\end{equation}
Each skill is first screened by its accumulated retry count $r_{\phi}$, which indicates where the standing guidance may not sufficiently cover the encountered task conditions. The skill with the highest retry count is selected for further inspection:
\begin{equation}
    \phi^{\star} = \arg\max_{\phi} r_{\phi}.
\end{equation}
The Curator then analyses its complete evidence record, including failed validation conditions, the recovery sequence, and the successfully adapted script, to identify the underlying deficiency and determine whether and how the skill should be updated.

\begin{figure}[!htbp]
    \centering
    \includegraphics[width=0.9\linewidth]{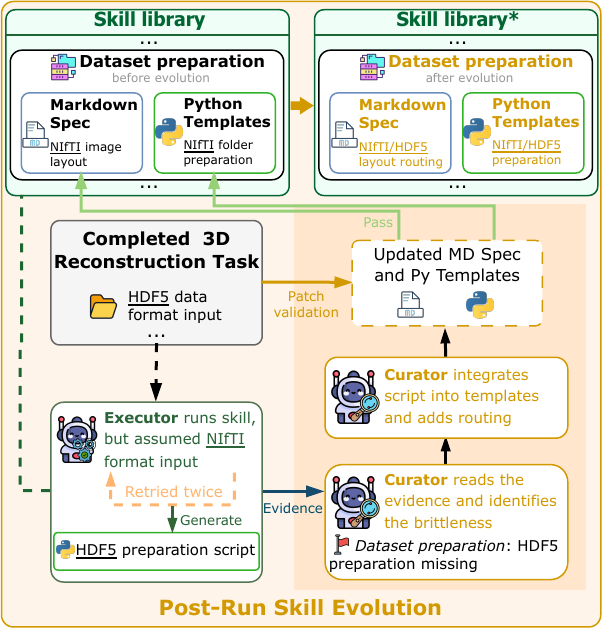}
    \caption{The Curator's post-run skill evolution loop after a 3D reconstruction task with HDF5 dataset format.}
    \label{fig:evolution}
    %\vspace{-8pt}
\end{figure}

As shown in Figure~\ref{fig:evolution}, a 3D reconstruction dataset supplied in HDF5 format forces the \textit{Dataset preparation} skill into repeated recovery, as the Executor initially assumes the NIfTI layout used for segmentation and adapts a new HDF5 preparation script only after retrying. The Curator aggregates the traces of this run into the skill's evidence record $\mathcal{E}_\phi$, capturing its elevated retry count, the validation logs of the failed format checks, and the adapted HDF5 script that ultimately succeeded. Brittleness identification across all skills then identifies \textit{Dataset preparation} as the brittleness maximiser $\phi^\star$, marking it as the candidate for patching.

\subsubsection{Patch Proposal and Validation}
For the flagged skill $\phi^\star$, the Curator analyses its evidence record $\mathcal{E}_{\phi^\star}$ to diagnose the deficiency underlying the retries, and proposes a patch $\delta$ comprising two complementary parts. 
1) The adapted script $\tau_{\phi^\star}^\star$ that the Executor produced to resolve the difficult case during the run, recorded in the evidence but not yet part of the library, is consolidated into the skill as a new reusable template; 
and 2) the skill's durable knowledge, its Markdown specification or reference document, is extended to describe the condition and when that template applies. 
Consolidating both the recovered script and the guidance that selects it transfers the handling of the condition from reactive recovery to proactive capability, generalising it to subsequent encounters rather than remedying only the instance that elicited it.

Before it is committed, the patch is validated by re-executing the affected step on the precise input $\mathcal{A}^\star$ that triggered the retries, and is accepted only if the re-execution satisfies the corresponding gate:
\begin{equation}
\phi^\star \leftarrow
\begin{cases}
\phi^\star \cup \delta & \text{if } g(\Phi(\mathcal{A}^\star;\, \phi^\star \cup \delta)) = 1 \\
\phi^\star & \text{otherwise.}
\end{cases}
\end{equation}
Validation requires the patched skill to pass the gate on its first execution without retries, indicating that the guidance now covers the condition rather than leaving it to reactive recovery.
A verified patch is committed additively, as $\phi^\star \cup \delta$, extending the skill while preserving previously covered cases, so evolution maintains coverage monotonically. 
Over successive runs, the library thus covers more of the encountered task distribution.

Figure~\ref{fig:evolution} depicts a representative instance for the \textit{Dataset preparation} skill: the Curator consolidates the recovered HDF5 preparation script into the skill's templates and extends its specification with a data-type routing clause that recognises HDF5 inputs up front, then validates the patch by re-executing preparation on the same dataset. Upon passing the gate, the patch is committed to the evolved skill library so that the recovered handling is applied proactively thereafter.

% -------------------------------------------------------------
\section{Evaluation}
\label{sec:evaluation}
We evaluate the proposed framework through a series of experiments designed to assess its end-to-end effectiveness, the contribution of its core design components, and its generalisability. Specifically, we investigate the following research questions:

\textbf{RQ1: How does EvoML compare against existing agentic AutoML systems and command-line coding tools on end-to-end ML pipeline synthesis?} \\
\textbf{RQ2: To what extent does validation-gated execution contribute to pipeline automation and model performance?}\\
\textbf{RQ3: How do expert grounding and post-run skill evolution contribute to pipeline synthesis performance?}\\
\textbf{RQ4: How well does the framework generalise beyond the targeted benchmarks to broader ML tasks?}

These RQs follow the design of EvoML and progressively examine its effectiveness. RQ1 evaluates the complete framework against existing agentic AutoML systems and coding tools. RQ2 isolates the contribution of validation-gated execution, while RQ3 examines the effects of expert grounding and post-run skill evolution. RQ4 then evaluates whether the framework extends beyond the primary medical-imaging setting to broader ML tasks.

\subsection{Benchmarking Datasets}
We evaluate primarily on medical imaging, a demanding testbed for ML pipeline synthesis. 
Existing ML-engineering benchmarks such as MLE-bench~\cite{chan2025mle} centre on competition-style tasks in which the challenge is maximising performance through model selection and tuning, rather than constructing a correct pipeline. 
Medical imaging instead combines heterogeneous formats, 2D and 3D volumes, and heavy modality-specific preprocessing, all under strict correctness constraints, yet a runnable pipeline can still be silently wrong. 
We select eight medical datasets spanning six task types:
\textbf{BTCV}~\cite{landman2015miccai}: supervised 3D segmentation of 13 organs in CT (50 cases; $512\times512\times123$), with few, imbalanced samples. 
\textbf{ACDC}~\cite{bernard2018deep}: supervised 3D cardiac segmentation in MRI ($215\times228\times10$), with pronounced voxel anisotropy. 
\textbf{FLARE21}~\cite{ma2022fast}: supervised 3D segmentation of 4 abdominal organs in CT (361 cases; $512\times512\times126$). 
\textbf{FLARE22}~\cite{ma2024unleashing}: semi-supervised 3D segmentation of 13 abdominal organs in CT (50 labelled, 2000 unlabelled; $512\times512\times196$), where label scarcity and semi-supervised training policy are the central difficulty. 
\textbf{Promise12MSBench}~\cite{kucs2024medsegbench}: supervised 2D binary prostate segmentation in MRI ($512\times512$), adapted from PROMISE12~\cite{litjens2014evaluation}.
\textbf{OrganMNIST3D}~\cite{yang2023medmnist}: supervised 3D classification of 11 organ categories in CT (1{,}742 samples; $64\times64\times64$), the least demanding task. 
\textbf{FastMRI}~\cite{zbontar2018fastmri}: supervised 3D reconstruction from undersampled k-space (60 cases; $288\times640\times16\times83$), demanding for its large HDF5 footprint and k-space pairing. 
\textbf{IXI}~\cite{ixi}: supervised 3D denoising of brain MRI (581 T1 volumes; $256\times256\times150$), requiring fine structure to be retained while suppressing noise.

\subsection{Evaluation Metrics}
We evaluate each system along two dimensions: its ability to autonomously construct an executable end-to-end ML pipeline, and the predictive quality of the resulting model. 
Accordingly, we report \emph{End-to-End Automation Capability (EAC)} and
\emph{Final Model Quality (FMQ)}.

\textbf{End-to-End Automation Capability (EAC).}
EAC measures how autonomously a method produces a complete, executable pipeline, penalised per step by the human effort required. We track five milestones: (1) dataset preparation, (2) model design, (3) training, (4) hyperparameter optimisation, and (5) pipeline integration. For milestone $k$ in run $r$, let $c_{r,k} \in \{0,1\}$ indicate completion with a valid artifact, and $h_{r,k}$ the number of interventions at that milestone. The per-milestone score is:
\begin{equation}
s_{r,k} = c_{r,k} \cdot \left(1 - \min\!\left(\frac{h_{r,k}}{3},\, 1\right)\right),
\end{equation}
where the intervention budget of three distinguishes minor recoverable failures from cases requiring substantial manual guidance. EAC is the mean across milestones:
\begin{equation}
\mathrm{EAC}_r = \frac{1}{5}\sum_{k=1}^{5} s_{r,k}.
\end{equation}
Each milestone contributes only when completed with a valid artifact, and its contribution decreases with the number of interventions it required.

% \textbf{Code Quality Score (CQS).}
% After a pipeline is generated, we evaluate the quality of the generated code using CQS. 
% This metric is motivated by the maintainability characteristic in the ISO/IEC 25010 software product quality model and is
% operationalised using measurable code-level indicators: maintainability index, cyclomatic complexity, and code duplication.

% For each run $r$, let $\mathrm{MI}_r\in[0,100]$ denote the Maintainability Index of the generated code, and let $\overline{\mathrm{CC}}_r$ denote the average cyclomatic complexity over all functions and methods, both computed using Radon~\cite{rubikrad2:online}. 
% Let $\mathrm{Dup}_r\in[0,1]$ denote the duplicated-line ratio computed using jscpd~\cite{kucheren65:online}, where duplicated lines refer to repeated or near-repeated code fragments within or across generated files. 
% We define the normalised maintainability, complexity, and duplication scores as $M_r=\mathrm{MI}_r/100$, $C_r=1-\min(\overline{\mathrm{CC}}_r/\tau_c,1)$, and $D_r=1-\mathrm{Dup}_r$, respectively, where $\tau_c=10$ follows the commonly used McCabe-style cyclomatic-complexity
% interpretation~\cite{watson1996structured} and matches Radon's complexity grading scheme.

% The final CQS is computed as:
% \begin{equation}
% \mathrm{CQS}_r
% =
% \frac{1}{3}
% \left(
% M_r+C_r+D_r
% \right).
% \end{equation}

\textbf{Final Model Quality (FMQ).}
FMQ measures the predictive performance of the final model. For segmentation
datasets, given predicted foreground mask $\hat{Y}_{r,i}$ and ground-truth mask
$Y_i$ for test case $i\in T$, we use the Dice score~\cite{zijdenbos1994morphometric,bertels2019optimizing}:
\begin{equation}
\mathrm{FMQ}_{\mathrm{seg},r}
=
\mathrm{Dice}_{r}
=
\frac{1}{|T|}
\sum_{i\in T}
\frac{
2|\hat{Y}_{r,i}\cap Y_i|
}{
|\hat{Y}_{r,i}|+|Y_i|
}.
\end{equation}

For classification datasets, we use accuracy:
\begin{equation}
\mathrm{FMQ}_{\mathrm{cls},r}
=
\mathrm{Acc}_{r}
=
\frac{1}{|T|}
\sum_{i\in T}
\mathbb{I}\!\left(\hat{y}_{r,i}=y_i\right),
\end{equation}
where $\hat{y}_{r,i}$ and $y_i$ denote the predicted and ground-truth labels,
respectively.

For the reconstruction task, we use the Structural Similarity Index Measure 
(SSIM)~\cite{wang2004image}:
\begin{equation}
\mathrm{FMQ}_{\mathrm{rec},r} = \mathrm{SSIM}_{r} =\frac{1}{|T|}
\sum_{i\in T}
\frac{(2\mu_p\mu_q+c_1)(2\sigma_{pq}+c_2)}
{(\mu_p^{2}+\mu_q^{2}+c_1)(\sigma_p^{2}+\sigma_q^{2}+c_2)}
,
\end{equation}
where $p=\hat{Y}_{r,i}$, $q=Y_i$, $\mu_p$, $\mu_q$ and $\sigma_p^2$, 
$\sigma_q^2$ denote the local means and variances, $\sigma_{pq}$ the local 
covariance, and $c_1$, $c_2$ are numerical stability constants.

For denoising, we use the Peak Signal-to-Noise Ratio (PSNR):
\begin{equation}
\mathrm{FMQ}_{\mathrm{den},r}
=
\mathrm{PSNR}_{r}
=
\frac{1}{|T|}
\sum_{i\in T}
10\log_{10}\!\left(
\frac{L^{2}}{\mathrm{MSE}(\hat{Y}_{r,i},\,Y_i)}
\right),
\end{equation}
where $L$ is the maximum possible pixel intensity value and
$\mathrm{MSE}(\hat{Y}_{r,i},Y_i)=\frac{1}{N}\sum_{j=1}^{N}
(\hat{Y}_{r,i,j}-Y_{i,j})^{2}$.

\subsection{Experiment Settings}
\subsubsection{Baseline Selection and Configuration}
We compare against two categories of baselines: three agentic AutoML systems targeting end-to-end ML pipeline construction, AutoML-Agent~\cite{trirat2024automl}, MLZero~\cite{fang2026mlzero}, and M3Builder~\cite{feng2025m}; and two widely adopted command-line coding tools, Codex~\cite{Introduc82:online} (GPT-5.5 Medium) and Claude Code~\cite{ClaudeCo70:online} (Opus 4.7 Medium).
To ensure a fair comparison, we apply the same task requirements to all baselines. 
For the agentic AutoML systems, we preserve each system's native prompting conventions so that it operates under the input format for which it was designed, and additionally supply the skeleton and template files that certain baselines require, following the setup specified in their respective implementations. For the command-line coding tools, we provide the same task description as EvoML and guide them through the same four-phase workflow, ensuring that all methods address an equivalent synthesis task.

\subsubsection{Environment and Protocol}
All experiments are conducted on an Ubuntu 22.04 server equipped with an AMD Ryzen Threadripper PRO 3995WX CPU (64 cores), 252GB of RAM, and two NVIDIA RTX A6000 GPUs (48GB VRAM each). 
We provide all methods in a unified Python environment pre-installed with ML and medical imaging libraries and common dependencies, and each method may install additional dependencies as needed during pipeline execution to ensure a fair comparison. 
To account for the stochasticity of LLM-based generation, each method is executed three times per dataset, and we report the mean $\pm$ std of EAC and FMQ.

\begin{figure*}[htbp]
\centering

\begin{subfigure}{\textwidth}
\centering
\begin{tikzpicture}
\begin{axis}[
    width=\textwidth, height=4cm,
    ybar=1pt, bar width=5pt,
    ymin=0.4, ymax=1.1,
    ylabel={\textbf{EAC}},
    symbolic x coords={BTCV,ACDC,FLARE21,FLARE22,Promise12,OrganMNIST,FastMRI,IXI},
    xtick=data, x tick label style={font=\scriptsize, rotate=20, anchor=east},
    ytick={0.4,0.5,0.6,0.7,0.8,0.9,1.0}, ylabel near ticks,
    legend style={at={(0.5,1.15)}, anchor=south, legend columns=6, font=\scriptsize},
    error bars/y dir=both, error bars/y explicit,
    enlarge x limits=0.06,
]
\addplot+[teal,fill=teal] coordinates {
  (BTCV,1)+-(0,0) (ACDC,1)+-(0,0) (FLARE21,1)+-(0,0) (FLARE22,1)+-(0,0) (Promise12,1)+-(0,0) (OrganMNIST,1)+-(0,0) (FastMRI,1)+-(0,0) (IXI,1)+-(0,0) };
\addplot+[amagent,fill=amagent] coordinates {
  (BTCV,0.7778)+-(0,0.1133) (ACDC,1)+-(0,0) (FLARE21,0.9555)+-(0,0.0314) (FLARE22,0.9333)+-(0,0) (Promise12,1)+-(0,0) (OrganMNIST,1)+-(0,0) (FastMRI,0.8222)+-(0,0.0314) (IXI,0.7555)+-(0,0.0314) };
\addplot+[orange,fill=orange] coordinates {
  (BTCV,0.7556)+-(0,0.0628) (ACDC,0.7778)+-(0,0.0314) (FLARE21,0.7333)+-(0,0.0544) (FLARE22,0.5111)+-(0,0.0314) (Promise12,0.7333)+-(0,0.0544) (OrganMNIST,0.7778)+-(0,0.0314) (FastMRI,0.7333)+-(0,0.0544) (IXI,0.7778)+-(0,0.0314) };
\addplot+[brown,fill=brown] coordinates {
  (BTCV,0.7333)+-(0,0.0544) (ACDC,0.6667)+-(0,0.0544) (FLARE21,0.6667)+-(0,0.0544) (FLARE22,0.4667)+-(0,0.0544) (Promise12,0.4667)+-(0,0.0544) (OrganMNIST,0.7333)+-(0,0.0544) (FastMRI,0.4667)+-(0,0.0544) (IXI,0.5333)+-(0,0.0544) };
\addplot+[codexcol,fill=codexcol] coordinates {
  (BTCV,0.9333)+-(0,0.0544) (ACDC,0.9111)+-(0,0.0314) (FLARE21,0.9111)+-(0,0.0314) (FLARE22,0.8445)+-(0,0.0314) (Promise12,0.8889)+-(0,0.0314) (OrganMNIST,0.8000)+-(0,0.0545) (FastMRI,0.8222)+-(0,0.0314) (IXI,0.7778)+-(0,0.0629) };
\addplot+[ccodecol,fill=ccodecol] coordinates {
  (BTCV,0.9778)+-(0,0.0314) (ACDC,0.9333)+-(0,0.0544) (FLARE21,0.9778)+-(0,0.0314) (FLARE22,0.9555)+-(0,0.0314) (Promise12,0.9333)+-(0,0) (OrganMNIST,1.0000)+-(0,0) (FastMRI,0.8000)+-(0,0.0943) (IXI,0.6667)+-(0,0) };
\legend{EvoML, AutoML-Agent, MLZero, M3Builder, Codex, Claude Code}
\end{axis}
\end{tikzpicture}
% \caption{EAC across the benchmarking datasets.}
\label{fig:rq1_eac}
\end{subfigure}

%\vspace{-5pt}

\begin{subfigure}{\textwidth}
\centering
\begin{tikzpicture}
\begin{axis}[
    width=\textwidth, height=4cm,
    ybar=1pt, bar width=5pt,
    ymin=0, ymax=1.1,
    ylabel={\textbf{FMQ}},
    symbolic x coords={BTCV,ACDC,FLARE21,FLARE22,Promise12,OrganMNIST,FastMRI,IXI},
    xtick=data, x tick label style={font=\scriptsize, rotate=20, anchor=east},
    ytick={0,0.2,0.4,0.6,0.8,1.0}, ylabel near ticks,
    error bars/y dir=both, error bars/y explicit,
    enlarge x limits=0.06,
]
\addplot+[teal,fill=teal] coordinates {
  (BTCV,0.8370)+-(0,0.0131) (ACDC,0.9186)+-(0,0.0043) (FLARE21,0.9353)+-(0,0.0022) (FLARE22,0.9199)+-(0,0.0060) (Promise12,0.9196)+-(0,0.0029) (OrganMNIST,0.9917)+-(0,0.0077) (FastMRI,0.9498)+-(0,0.0025) (IXI,0.7859)+-(0,0.0067) };
\addplot+[amagent,fill=amagent] coordinates {
  (BTCV,0.7170)+-(0,0.0549) (ACDC,0.3837)+-(0,0.1191) (FLARE21,0.8608)+-(0,0.0185) (FLARE22,0.5664)+-(0,0.0549) (Promise12,0.7802)+-(0,0.0279) (OrganMNIST,0.9876)+-(0,0.0088) (FastMRI,0.6583)+-(0,0.0164) (IXI,0.4405)+-(0,0.2127) };
\addplot+[orange,fill=orange] coordinates {
  (BTCV,0.5937)+-(0,0.0852) (ACDC,0.8652)+-(0,0.0145) (FLARE21,0.8478)+-(0,0.0282) (FLARE22,0.6343)+-(0,0.0425) (Promise12,0.7587)+-(0,0.0439) (OrganMNIST,0.8975)+-(0,0.0690) (FastMRI,0.6953)+-(0,0.0119) (IXI,0.7215)+-(0,0.0182) };
\addplot+[brown,fill=brown] coordinates {
  (BTCV,0.7982)+-(0,0.0060) (ACDC,0.9079)+-(0,0.0106) (FLARE21,0.9289)+-(0,0.0144) (FLARE22,0.5981)+-(0,0.1317) (Promise12,0.6439)+-(0,0.1287) (OrganMNIST,0.9109)+-(0,0.0016) (FastMRI,0.1979)+-(0,0.0005) (IXI,0.6859)+-(0,0.0002) };
\addplot+[codexcol,fill=codexcol] coordinates {
  (BTCV,0.6167)+-(0,0.0132) (ACDC,0.8628)+-(0,0.0176) (FLARE21,0.8416)+-(0,0.0221) (FLARE22,0.8532)+-(0,0.0157) (Promise12,0.9004)+-(0,0.0101) (OrganMNIST,0.9767)+-(0,0.0105) (FastMRI,0.8579)+-(0,0.0065) (IXI,0.5309)+-(0,0.0306) };
\addplot+[ccodecol,fill=ccodecol] coordinates {
  (BTCV,0.8307)+-(0,0.0118) (ACDC,0.7985)+-(0,0.0110) (FLARE21,0.9175)+-(0,0.0065) (FLARE22,0.8824)+-(0,0.0063) (Promise12,0.9033)+-(0,0.0055) (OrganMNIST,0.9311)+-(0,0.0084) (FastMRI,0.8521)+-(0,0.0097) (IXI,0.7064)+-(0,0.0118) };
\end{axis}
\end{tikzpicture}
% \caption{FMQ across the benchmarking datasets.}
\label{fig:rq1_fmq}
\end{subfigure}
%\vspace{-20pt}
\caption{Comparison of EvoML against agentic AutoML systems and command-line coding tools across the benchmarking datasets: EvoML attains full automation and the best FMQ on every dataset, with markedly lower variance than the baselines.}
\label{fig:rq1}
%\vspace{-8pt}
\end{figure*}
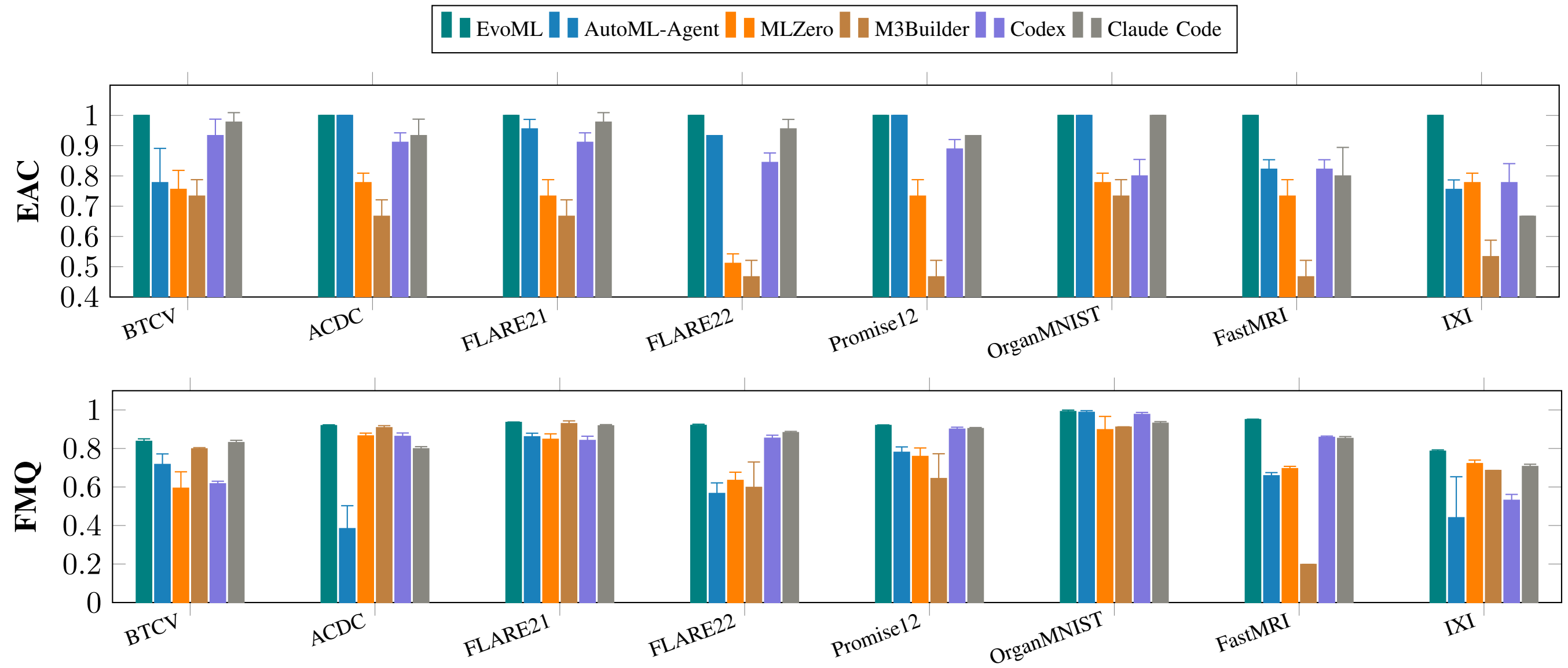

\subsection{Experiment Results}
\subsubsection{Comparison with Baseline Methods}
Figure~\ref{fig:rq1} compares EvoML against the five baselines across the eight datasets.
On automation (EAC), EvoML is the only method that reaches $1.0$ on every dataset, as its validation-gated harness resolves stage failures within the retry budget without human intervention. The baselines reflect their design assumptions. M3Builder handles supervised 3D segmentation reliably but falls short on the semi-supervised and non-segmentation tasks, where its EAC drops to $0.4667$ on FLARE22 and FastMRI. MLZero, relying on documentation retrieval and iterative trial rather than a fixed workflow, completes pipelines unevenly ($0.5111$ to $0.7778$) as its exploratory loops stall on harder tasks. AutoML-Agent, lifted by the code skeleton we supply, reaches near $1.0$ on the easier segmentation and classification tasks but fails to adapt to harder scenarios, dropping to $0.7778$ on BTCV and $0.7555$ on IXI. The command-line tools are the most autonomous baselines, recovering from runtime errors by monitoring the terminal, yet they require intervention on the reconstruction and denoising tasks, where failures arise from modelling design rather than execution.

On model performance (FMQ), EvoML attains the best result on every dataset, owing to its expert-grounded skills and the systematic HPO and promotion the baselines lack. The margin is largest on the harder tasks, reaching $0.9498$ on FastMRI against the next-best $0.8579$ and $0.7859$ on IXI against $0.7215$. The baselines degrade in ways that trace to their design: the command-line tools stay competitive on easier tasks but fall away on the hardest, with Codex dropping to $0.6167$ on BTCV and $0.5309$ on IXI, while the dedicated AutoML systems collapse more sharply, AutoML-Agent to $0.3837$ on ACDC and M3Builder to $0.1979$ on FastMRI.

\begin{table}[!htbp]
\centering
\caption{FastMRI reconstruction pipeline design choices across methods, with resulting mean FMQ (SSIM).}
\label{tab:fastmri_design}
\setlength{\tabcolsep}{4pt}
\renewcommand{\arraystretch}{1.2}
\begin{tabularx}{\columnwidth}{@{}l X X X c@{}}
\toprule
Method & Input Pairing & Model & Loss / Output & FMQ \\
\midrule
Codex & Synthesised from target & U-Net3D & L1+SSIM, 1-ch & 0.8579 \\
Claude Code & Synthesised from target & U-Net3D & L1+SSIM, 1-ch & 0.8521 \\
MLZero & Backprojection $\to$ temptv & U-Net2D & L1, 1-ch & 0.6953 \\
AutoML-Agent & Zero-filled $\to$ temptv & U-Net2D & SSIM, 4-ch sigmoid & 0.6583 \\
M3Builder & Zero-filled $\to$ temptv & U-Net3DTiny & L1+SSIM, 4-ch sigmoid & 0.1979 \\
\midrule
EvoML & Undersampled $\to$ full k-space RSS & VarNet3D & L1+SSIM, 1-ch & \textbf{0.9498} \\
\bottomrule
\end{tabularx}
%\vspace{-5pt}
\end{table}
We take FastMRI as a representative case, since reconstruction exposes the widest range of quality and allows clear right-and-wrong design choices. 
Table~\ref{tab:fastmri_design} compares each method's input--target pairing strategy, model design, and loss function selection, and the results track how faithfully each handles the reconstruction-specific requirements. 
The command-line tools adopt a sound model and loss function but shortcut the data pairing, synthesising the input from the target rather than the measured k-space, which simplifies the task and yields competitive but lower SSIM. 
MLZero makes sounder choices, reconstructing via backprojection with an L1-trained 2D U-Net. 
AutoML-Agent and M$^3$Builder fare the worst, importing segmentation defaults unsuited to reconstruction: both pair a coarsely resized, zero-filled reconstruction with a mismatched target and retain a multi-channel sigmoid output suited to label maps rather than intensity regression, with M$^3$Builder collapsing to $0.1979$ under the most aggressive resizing. 
EvoML follows the correct protocol, undersampling the fully sampled k-space relative to the full RSS target, using a reconstruction-specific VarNet3D and an L1--SSIM objective, achieving the highest FMQ. Final quality thus depends less on model capacity than on whether the pipeline respects the task structure, which is what EvoML's expert grounding enforces.

EvoML is also markedly more consistent, with FMQ standard deviation below $0.013$ on every dataset, whereas the baselines fluctuate widely (AutoML-Agent $\pm0.2127$ on IXI, M3Builder $\pm0.1317$ on FLARE22). This stability stems from the expert-grounded skills, which anchor every run in the same proven nnU-Net procedure rather than re-deriving the pipeline, while evolution reinforces it by settling the correct task-specific handling into the skills so quality-determining choices are applied reliably across runs.
%\vspace{-5pt}
\begin{tcolorbox}[colback=gray!5,colframe=gray!50,boxrule=0.5pt,arc=10pt,left=2pt,right=2pt,top=2pt,bottom=2pt]
\textbf{Answer to RQ1:} 
EvoML outperforms all five baselines, achieving full automation and the best final model quality on every dataset with markedly lower variance. This stems from three complementary strengths: expert-grounded skills produce consistently strong models, post-run evolution stabilises quality across runs, and the validation-gated harness sustains full autonomy.
% The baselines either complete pipelines but produce lower-quality models, as with the command-line coding tools, or fail to adapt to the full range of tasks, as with the dedicated AutoML systems.
\end{tcolorbox}

\subsubsection{Ablation Study}
We conduct two ablations to isolate the contributions of EvoML's core components: the validation-gated execution harness and the expert-grounded, self-evolving skill design. Each ablation removes or varies one component while holding the rest of the framework fixed, and is evaluated under the same protocol as the main comparison.

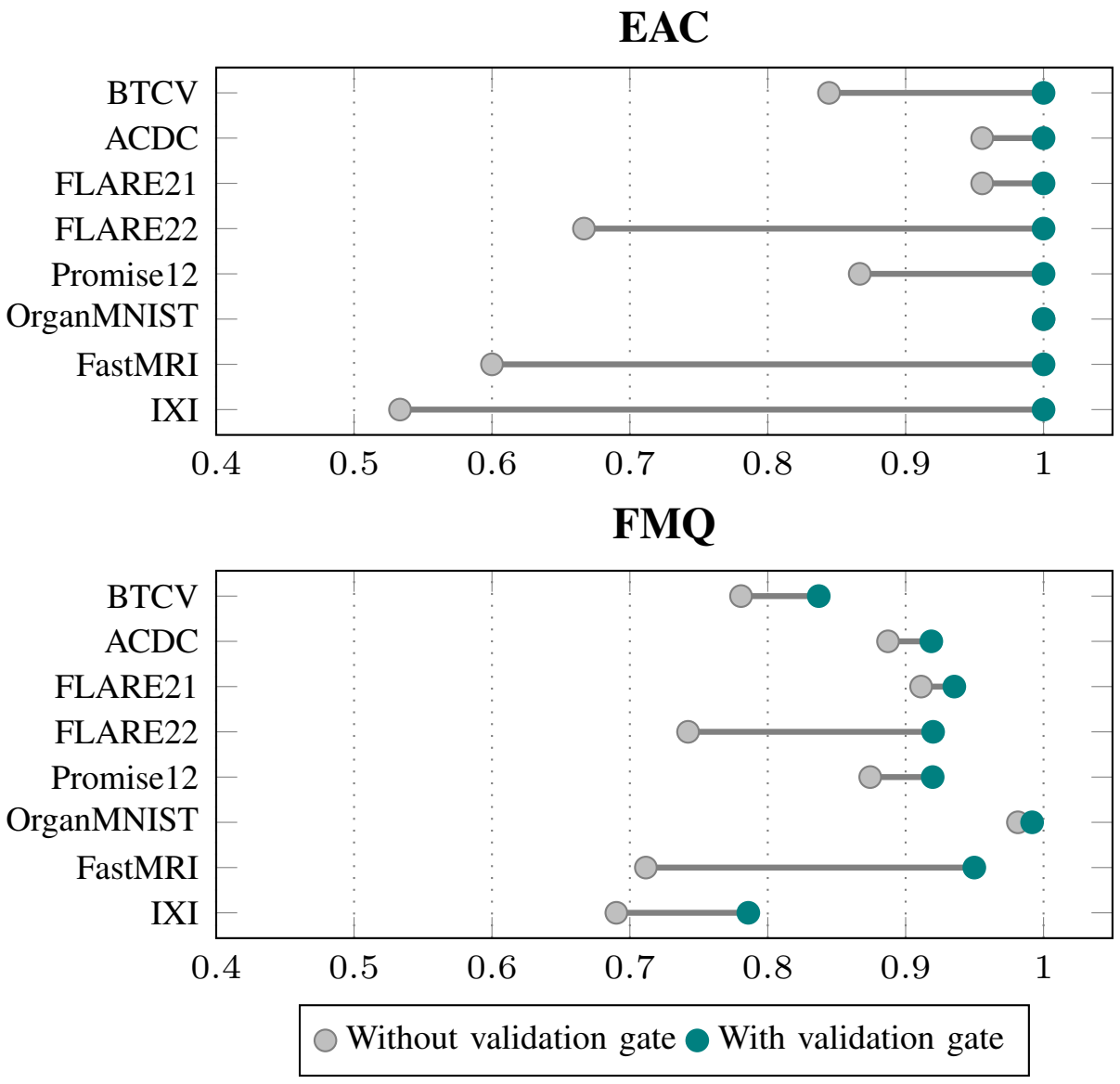
\begin{figure}[!htbp]
\centering

\begin{subfigure}{\columnwidth}
\centering
\begin{tikzpicture}
\begin{axis}[
    width=0.9\columnwidth, height=4.2cm,
    xmin=0.4, xmax=1.05,
    ytick={0,1,2,3,4,5,6,7},
    yticklabels={IXI,FastMRI,OrganMNIST,Promise12,FLARE22,FLARE21,ACDC,BTCV},
    yticklabel style={font=\scriptsize},
    xticklabel style={font=\scriptsize},
    xmajorgrids, grid style={dotted,gray},
    enlarge y limits=0.08,
    title={\textbf{EAC}}, title style={font=\small, yshift=-1ex},
]
% connectors: \draw (no-gate) -- (full) for each dataset
% order bottom->top: IXI,FastMRI,OrganMNIST,Promise12,FLARE22,FLARE21,ACDC,BTCV
\draw[gray,line width=1.2pt] (axis cs:0.5333,0)--(axis cs:1.0,0);
\draw[gray,line width=1.2pt] (axis cs:0.6000,1)--(axis cs:1.0,1);
\draw[gray,line width=1.2pt] (axis cs:1.0000,2)--(axis cs:1.0,2);
\draw[gray,line width=1.2pt] (axis cs:0.8667,3)--(axis cs:1.0,3);
\draw[gray,line width=1.2pt] (axis cs:0.6667,4)--(axis cs:1.0,4);
\draw[gray,line width=1.2pt] (axis cs:0.9555,5)--(axis cs:1.0,5);
\draw[gray,line width=1.2pt] (axis cs:0.9555,6)--(axis cs:1.0,6);
\draw[gray,line width=1.2pt] (axis cs:0.8444,7)--(axis cs:1.0,7);
% no-gate points (open/grey)
\addplot[only marks,mark=*,mark options={fill=gray!50,draw=gray},mark size=2.2pt] coordinates {
 (0.5333,0)(0.6000,1)(1.0000,2)(0.8667,3)(0.6667,4)(0.9555,5)(0.9555,6)(0.8444,7)};
% full points (teal)
\addplot[only marks,mark=*,mark options={fill=teal,draw=teal},mark size=2.2pt] coordinates {
 (1.0,0)(1.0,1)(1.0,2)(1.0,3)(1.0,4)(1.0,5)(1.0,6)(1.0,7)};
\end{axis}
\end{tikzpicture}
% \caption{EAC}
\label{fig:rq2_eac}
\end{subfigure}

% %\vspace{-3pt}

\begin{subfigure}{\columnwidth}
\centering
\begin{tikzpicture}
\begin{axis}[
    width=0.9\columnwidth, height=4.2cm,
    xmin=0.4, xmax=1.05,
    ytick={0,1,2,3,4,5,6,7},
    yticklabels={IXI,FastMRI,OrganMNIST,Promise12,FLARE22,FLARE21,ACDC,BTCV},
    yticklabel style={font=\scriptsize},
    xticklabel style={font=\scriptsize},
    xmajorgrids, grid style={dotted,gray},
    enlarge y limits=0.08,
    title={\textbf{FMQ}}, title style={font=\small, yshift=-1ex},
    legend style={at={(0.5,-0.18)},anchor=north,legend columns=2,font=\scriptsize},
]
\draw[gray,line width=1.2pt] (axis cs:0.6902,0)--(axis cs:0.7859,0);
\draw[gray,line width=1.2pt] (axis cs:0.7116,1)--(axis cs:0.9498,1);
\draw[gray,line width=1.2pt] (axis cs:0.9814,2)--(axis cs:0.9917,2);
\draw[gray,line width=1.2pt] (axis cs:0.8743,3)--(axis cs:0.9196,3);
\draw[gray,line width=1.2pt] (axis cs:0.7422,4)--(axis cs:0.9199,4);
\draw[gray,line width=1.2pt] (axis cs:0.9112,5)--(axis cs:0.9353,5);
\draw[gray,line width=1.2pt] (axis cs:0.8872,6)--(axis cs:0.9186,6);
\draw[gray,line width=1.2pt] (axis cs:0.7806,7)--(axis cs:0.8370,7);
\addplot[only marks,mark=*,mark options={fill=gray!50,draw=gray},mark size=2.2pt] coordinates {
 (0.6902,0)(0.7116,1)(0.9814,2)(0.8743,3)(0.7422,4)(0.9112,5)(0.8872,6)(0.7806,7)};
\addlegendentry{Without validation gate}
\addplot[only marks,mark=*,mark options={fill=teal,draw=teal},mark size=2.2pt] coordinates {
 (0.7859,0)(0.9498,1)(0.9917,2)(0.9196,3)(0.9199,4)(0.9353,5)(0.9186,6)(0.8370,7)};
\addlegendentry{With validation gate}
\end{axis}
\end{tikzpicture}
% \caption{FMQ}
\label{fig:rq2_fmq}
\end{subfigure}

\caption{Effect of validation gating: without gates, EAC and FMQ both decline, sharply on the harder reconstruction, denoising, and semi-supervised tasks.}
\label{fig:rq2}
\end{figure}

\textbf{Validation Gating.} To isolate the contribution of validation-gated execution, we compare full EvoML against a variant in which the gates are removed, so stages progress unconditionally and failures are neither detected nor automatically recovered. 
Figure~\ref{fig:rq2} reports both metrics across the eight datasets. The removal of validation gates degrades automation wherever failures arise: a faulty artifact can no longer be detected and repaired in place, so pipeline completion reverts to manual intervention, and EAC declines from $1.0$ to as low as $0.5333$ on IXI and $0.6000$ on FastMRI. This degradation scales with task difficulty, leaving the simpler OrganMNIST, FLARE21, and ACDC largely unaffected while the more demanding reconstruction, denoising, and semi-supervised tasks deteriorate substantially.
Model quality declines in parallel, though for a more subtle reason. The errors intercepted by the gates are frequently not outright failures but silent inconsistencies, such as the application of an MRI-style normalisation to CT data, which execute without error yet corrupt the training process. Left unchecked, such errors propagate into the model and manifest only as degraded output: FMQ falls most markedly on the hardest tasks, declining from $0.9199$ to $0.7422$ on FLARE22 and from $0.9498$ to $0.7116$ on FastMRI, while the simplest task, OrganMNIST, remains nearly unaffected. The gates thus safeguard both autonomy, by recovering failures automatically, and quality, by arresting silent errors before they reach the model.
%\vspace{-5pt}
\begin{tcolorbox}[colback=gray!5,colframe=gray!50,boxrule=0.5pt,arc=10pt,left=2pt,right=2pt,top=2pt,bottom=2pt]
\textbf{Answer to RQ2:} Validation gating underpins both automation and quality: by detecting and repairing faulty artifacts in place, it sustains autonomy; by intercepting silent inconsistencies, it prevents the error propagation that would otherwise degrade model performance.
\end{tcolorbox}

\textbf{Skill Expert-Grounding and Evolution.} We examine the skill design along two axes: the contribution of expert grounding, and the contribution of post-run evolution.
\begin{figure}[!htbp]
\centering

\begin{subfigure}{\columnwidth}
\centering
\begin{tikzpicture}
\begin{axis}[
    width=\columnwidth, height=4cm,
    ybar=1.5pt, bar width=6pt,
    ymin=0.4, ymax=1.05,
    ylabel={\textbf{EAC}},
    symbolic x coords={BTCV,ACDC,FLARE21,FLARE22,Promise12,OrganMNIST,FastMRI,IXI},
    xtick=data, x tick label style={font=\scriptsize, rotate=20, anchor=east},
    ytick={0.4,0.5,0.6,0.7,0.8,0.9,1.0}, ylabel near ticks,
    legend style={at={(0.5,1.18)}, anchor=south, legend columns=2, font=\scriptsize},
    error bars/y dir=both, error bars/y explicit,
    enlarge x limits=0.06,
]
\addplot+[teal,fill=teal] coordinates {
  (BTCV,1)+-(0,0) (ACDC,1)+-(0,0) (FLARE21,1)+-(0,0) (FLARE22,1)+-(0,0) (Promise12,1)+-(0,0) (OrganMNIST,1)+-(0,0) (FastMRI,1)+-(0,0) (IXI,1)+-(0,0) };
\addplot+[gencol,fill=gencol] coordinates {
  (BTCV,0.7111)+-(0,0.0831) (ACDC,0.9333)+-(0,0.0544) (FLARE21,0.9555)+-(0,0.0314) (FLARE22,0.6667)+-(0,0.1089) (Promise12,0.8889)+-(0,0.0831) (OrganMNIST,1.0)+-(0,0) (FastMRI,0.7556)+-(0,0.0832) (IXI,0.7333)+-(0,0.0544) };
\legend{Expert-grounded, General coding skills}
\end{axis}
\end{tikzpicture}
% \caption{EAC}
\label{fig:rq3_ground_eac}
\end{subfigure}

\par%\vspace{1ex}

\begin{subfigure}{\columnwidth}
\centering
\begin{tikzpicture}
\begin{axis}[
    width=\columnwidth, height=4cm,
    ybar=1.5pt, bar width=6pt,
    ymin=0.5, ymax=1.05,
    ylabel={\textbf{FMQ}},
    symbolic x coords={BTCV,ACDC,FLARE21,FLARE22,Promise12,OrganMNIST,FastMRI,IXI},
    xtick=data, x tick label style={font=\scriptsize, rotate=20, anchor=east},
    ytick={0.5,0.6,0.7,0.8,0.9,1.0}, ylabel near ticks,
    error bars/y dir=both, error bars/y explicit,
    enlarge x limits=0.06,
]
\addplot+[teal,fill=teal] coordinates {
  (BTCV,0.8370)+-(0,0.0131) (ACDC,0.9186)+-(0,0.0043) (FLARE21,0.9353)+-(0,0.0022) (FLARE22,0.9199)+-(0,0.0060) (Promise12,0.9196)+-(0,0.0029) (OrganMNIST,0.9917)+-(0,0.0088) (FastMRI,0.9498)+-(0,0.0025) (IXI,0.7859)+-(0,0.0067) };
\addplot+[gencol,fill=gencol] coordinates {
  (BTCV,0.7003)+-(0,0.0256) (ACDC,0.8590)+-(0,0.0126) (FLARE21,0.8790)+-(0,0.0131) (FLARE22,0.7150)+-(0,0.0363) (Promise12,0.8190)+-(0,0.0230) (OrganMNIST,0.9807)+-(0,0.0069) (FastMRI,0.7967)+-(0,0.0290) (IXI,0.6933)+-(0,0.0205) };
\end{axis}
\end{tikzpicture}
% \caption{FMQ}
\label{fig:rq3_ground_fmq}
\end{subfigure}

\caption{Effect of expert grounding: without the nnU-Net grounding, both automation and quality fall and variance roughly doubles.}
\label{fig:rq3_ground}
%\vspace{-8pt}
\end{figure}
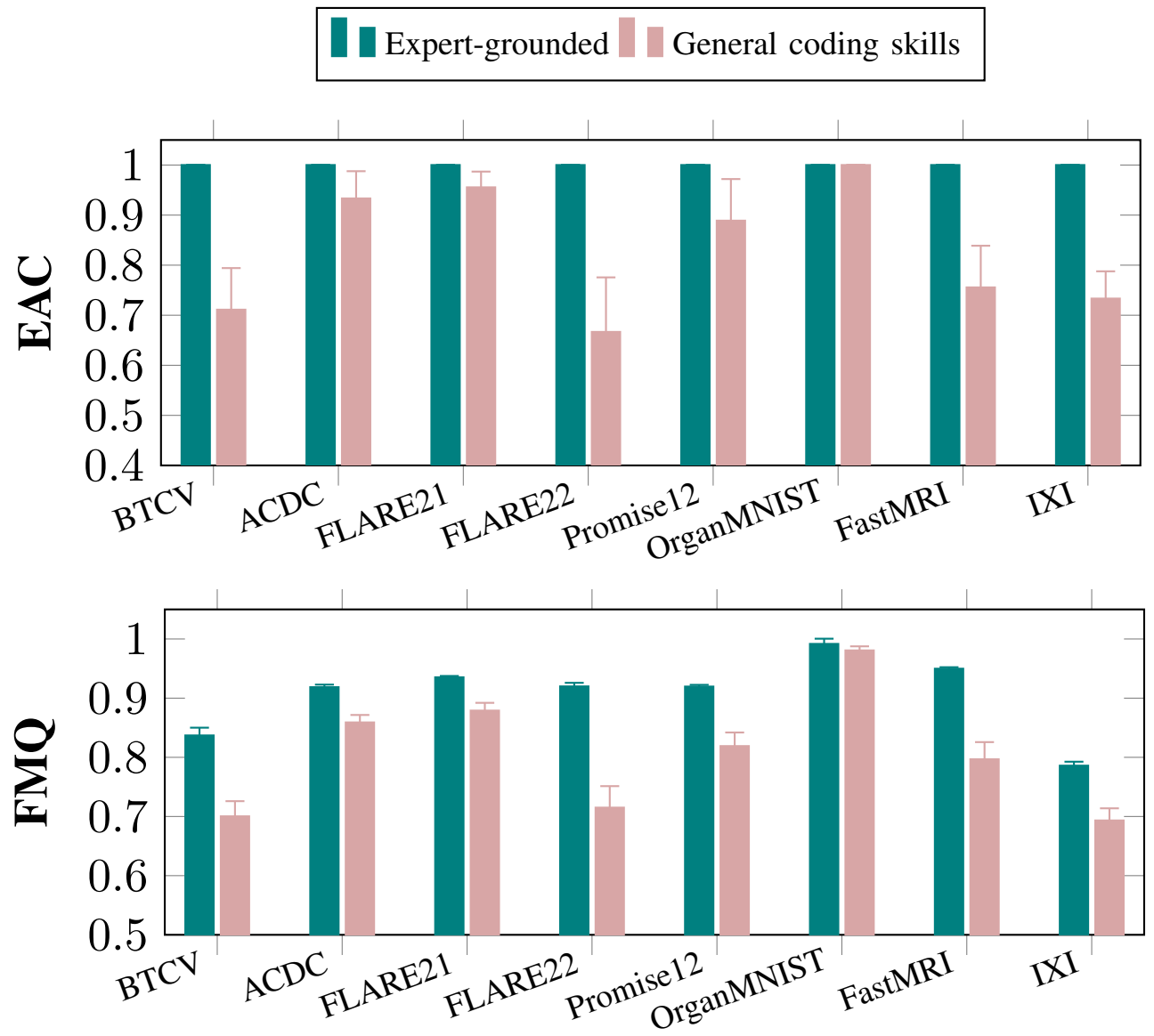

% \textit{Expert grounding.} We compare full EvoML against a variant that keeps the skill structure, harness, and evolution but removes the nnU-Net grounding, replacing dataset preparation, fingerprinting, planning, preprocessing, training, and evaluation with generic, executor-designed equivalents. As Figure~\ref{fig:rq3_ground} shows, the executor's ad hoc steps fail more often, lowering EAC to as low as $0.6667$ on FLARE22, and yield weaker models, with FMQ dropping from $0.9199$ to $0.7150$ on FLARE22 and $0.9498$ to $0.7967$ on FastMRI; run-to-run variance also roughly doubles. Expert grounding therefore raises quality, sustains autonomy, and reduces variance.
\textit{Expert-Grounding.} We compare full EvoML against a variant that keeps the skill structure, harness, and evolution but removes the nnU-Net grounding, replacing dataset preparation, fingerprinting, planning, preprocessing, training, and evaluation with generic, executor-designed equivalents. Lacking a proven procedure to follow, the executor re-derives each step from scratch, so its choices vary from run to run and often miss the dataset-specific handling the grounded skills encode. As Figure~\ref{fig:rq3_ground} shows, these ad hoc steps fail more often, lowering EAC to as low as $0.6667$ on FLARE22, and yield weaker models, with FMQ dropping from $0.9199$ to $0.7150$ on FLARE22 and $0.9498$ to $0.7967$ on FastMRI; run-to-run variance also roughly doubles, since each run reconstructs the pipeline independently. Expert grounding therefore raises quality, sustains autonomy, and reduces variance.

\definecolor{noevocol}{HTML}{BA7517}

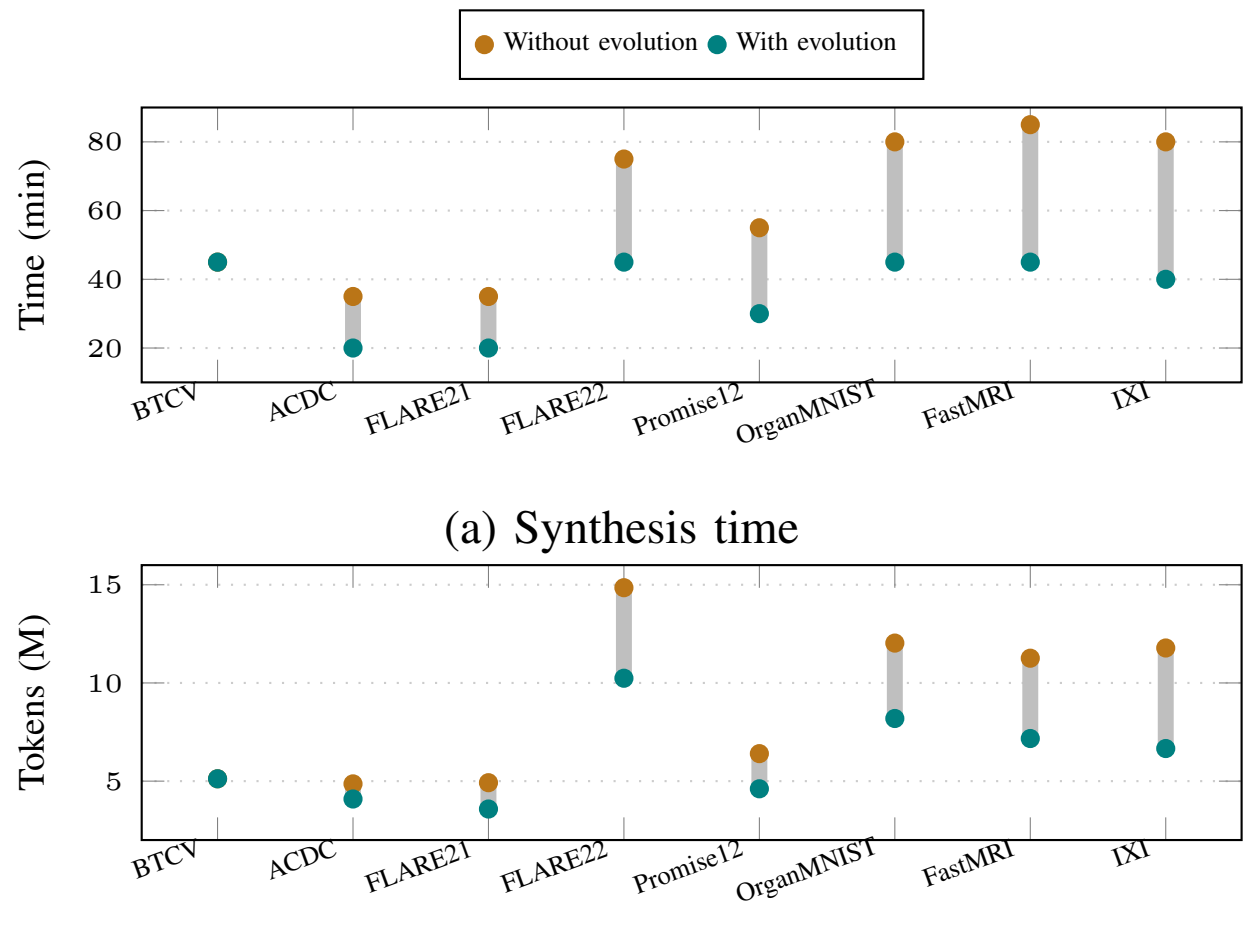
\begin{figure}[!htbp]
\centering
\pgfplotsset{
  evopanel/.style={
    width=\columnwidth, height=3.4cm,
    symbolic x coords={BTCV,ACDC,FLARE21,FLARE22,Promise12,OrganMNIST,FastMRI,IXI},
    xtick=data, x tick label style={font=\tiny, rotate=20, anchor=east},
    yticklabel style={font=\tiny}, ylabel style={font=\scriptsize},
    enlarge x limits=0.08, ymajorgrids, grid style={dotted,gray!40},
  }
}

% ---- Time ----
\begin{subfigure}{\columnwidth}
\centering
\begin{tikzpicture}
\begin{axis}[evopanel, ymin=10, ymax=90, ylabel={Time (min)}, legend style={at={(0.5,1.10)},anchor=south,legend columns=2,font=\tiny}]
\draw[gray!50,line width=3pt] (axis cs:ACDC,20)--(axis cs:ACDC,35);
\draw[gray!50,line width=3pt] (axis cs:FLARE21,20)--(axis cs:FLARE21,35);
\draw[gray!50,line width=3pt] (axis cs:FLARE22,45)--(axis cs:FLARE22,75);
\draw[gray!50,line width=3pt] (axis cs:Promise12,30)--(axis cs:Promise12,55);
\draw[gray!50,line width=3pt] (axis cs:OrganMNIST,45)--(axis cs:OrganMNIST,80);
\draw[gray!50,line width=3pt] (axis cs:FastMRI,45)--(axis cs:FastMRI,85);
\draw[gray!50,line width=3pt] (axis cs:IXI,40)--(axis cs:IXI,80);
\addplot[only marks,mark=*,noevocol,mark size=1.6pt] coordinates {(BTCV,45)(ACDC,35)(FLARE21,35)(FLARE22,75)(Promise12,55)(OrganMNIST,80)(FastMRI,85)(IXI,80)};
\addlegendentry{Without evolution}
\addplot[only marks,mark=*,teal,mark size=1.6pt] coordinates {(BTCV,45)(ACDC,20)(FLARE21,20)(FLARE22,45)(Promise12,30)(OrganMNIST,45)(FastMRI,45)(IXI,40)};
\addlegendentry{With evolution}
\end{axis}
\end{tikzpicture}
%\vspace{-8pt}
\caption{Synthesis time}
\label{fig:rq3_evo_time}
\end{subfigure}

% \par%\vspace{2ex}
% ---- Tokens ----
\begin{subfigure}{\columnwidth}
\centering
\begin{tikzpicture}
\begin{axis}[evopanel, ymin=2, ymax=16, ylabel={Tokens (M)}]
\draw[gray!50,line width=3pt] (axis cs:ACDC,4.09)--(axis cs:ACDC,4.86);
\draw[gray!50,line width=3pt] (axis cs:FLARE21,3.58)--(axis cs:FLARE21,4.92);
\draw[gray!50,line width=3pt] (axis cs:FLARE22,10.24)--(axis cs:FLARE22,14.85);
\draw[gray!50,line width=3pt] (axis cs:Promise12,4.61)--(axis cs:Promise12,6.40);
\draw[gray!50,line width=3pt] (axis cs:OrganMNIST,8.19)--(axis cs:OrganMNIST,12.03);
\draw[gray!50,line width=3pt] (axis cs:FastMRI,7.17)--(axis cs:FastMRI,11.26);
\draw[gray!50,line width=3pt] (axis cs:IXI,6.66)--(axis cs:IXI,11.78);
\addplot[only marks,mark=*,noevocol,mark size=1.6pt] coordinates {(BTCV,5.12)(ACDC,4.86)(FLARE21,4.92)(FLARE22,14.85)(Promise12,6.40)(OrganMNIST,12.03)(FastMRI,11.26)(IXI,11.78)};
\addplot[only marks,mark=*,teal,mark size=1.6pt] coordinates {(BTCV,5.12)(ACDC,4.09)(FLARE21,3.58)(FLARE22,10.24)(Promise12,4.61)(OrganMNIST,8.19)(FastMRI,7.17)(IXI,6.66)};
\end{axis}
\end{tikzpicture}
%\vspace{-8pt}
\caption{Token consumption}
\label{fig:rq3_evo_tok}
\end{subfigure}

% %\vspace{-2pt}

% ---- FMQ ----
% \begin{subfigure}{\columnwidth}
% \centering
% \begin{tikzpicture}
% \begin{axis}[evopanel, ymin=0.74, ymax=1.0, ylabel={FMQ}]
% \draw[gray!50,line width=3pt] (axis cs:BTCV,0.8363)--(axis cs:BTCV,0.8370);
% \draw[gray!50,line width=3pt] (axis cs:ACDC,0.9130)--(axis cs:ACDC,0.9186);
% \draw[gray!50,line width=3pt] (axis cs:FLARE21,0.9243)--(axis cs:FLARE21,0.9353);
% \draw[gray!50,line width=3pt] (axis cs:FLARE22,0.8843)--(axis cs:FLARE22,0.9199);
% \draw[gray!50,line width=3pt] (axis cs:Promise12,0.9033)--(axis cs:Promise12,0.9196);
% \draw[gray!50,line width=3pt] (axis cs:OrganMNIST,0.9823)--(axis cs:OrganMNIST,0.9917);
% \draw[gray!50,line width=3pt] (axis cs:FastMRI,0.9380)--(axis cs:FastMRI,0.9498);
% \draw[gray!50,line width=3pt] (axis cs:IXI,0.7573)--(axis cs:IXI,0.7859);
% \addplot[only marks,mark=*,noevocol,mark size=1.6pt] coordinates {(BTCV,0.8363)(ACDC,0.9130)(FLARE21,0.9243)(FLARE22,0.8843)(Promise12,0.9033)(OrganMNIST,0.9823)(FastMRI,0.9380)(IXI,0.7573)};
% \addplot[only marks,mark=*,teal,mark size=1.6pt] coordinates {(BTCV,0.8370)(ACDC,0.9186)(FLARE21,0.9353)(FLARE22,0.9199)(Promise12,0.9196)(OrganMNIST,0.9917)(FastMRI,0.9498)(IXI,0.7859)};

% \end{axis}
% \end{tikzpicture}
% %\vspace{-8pt}
% \caption{FMQ}
% \label{fig:rq3_evo_fmq}
% \end{subfigure}
\caption{Effect of post-run evolution: evolution reduces synthesis time and token consumption across the dataset sequence.}
\label{fig:rq3_evo}
%\vspace{-8pt}
\end{figure}

\textit{Skill Evolution.} We isolate post-run evolution by disabling the Curator: the Executor still adapts templates online, but neither the adapted scripts nor the guidance refinements are retained, so every dataset runs against the original library. 
Both configurations run over an identical, fixed dataset order, from segmentation (BTCV, ACDC, FLARE21, the semi-supervised FLARE22, Promise12) through classification (OrganMNIST) to the hardest reconstruction and denoising tasks (FastMRI, IXI).
Only the evolving configuration accumulates experience across this sequence.
Results are shown in Figure~\ref{fig:rq3_evo}. Evolution reduces total synthesis time by roughly 41\% (from 490 to 290 minutes) and token consumption by roughly 30\% (from 71 to 50 million), as later datasets can reuse previously recovered scripts and guidance instead of rediscovering them through repeated adaptation. Beyond efficiency, we assess evolution using three process-level criteria: \textit{first-pass skill success}, whether a skill completes without recovery; \textit{validation-triggered retries}, the number of recovery attempts caused by failed gates; and \textit{cross-task reuse}, whether adaptations retained from earlier datasets improve first-pass execution on later tasks. Excluding BTCV, where no prior evolution is available, first-pass skill success increases from 36/63 (57.1\%) to 50/63 (79.4\%), while validation-triggered retries decrease from 27 to 13 ($-51.9\%$). Cross-task reuse is also visible on later datasets: Promise12 and OrganMNIST complete all nine skill executions on the first pass, while IXI improves from 3/9 to 9/9 after accumulated evolution. The two configurations coincide on BTCV, confirming that these gains arise only after experience has accumulated across runs.
These process-level criteria complement FMQ by capturing changes in execution behaviour across runs, allowing us to distinguish retained procedural improvement from changes in final predictive performance.
%\vspace{-5pt}
\begin{tcolorbox}[colback=gray!5,colframe=gray!50,boxrule=0.5pt,arc=10pt,left=2pt,right=2pt,top=2pt,bottom=2pt]
\textbf{Answer to RQ3:} Expert grounding and post-run evolution contribute in complementary ways. Expert grounding improves model quality, autonomy, and consistency, while evolution reduces repeated recovery, improves first-pass execution, and lowers synthesis time and token consumption as experience accumulates across tasks.
\end{tcolorbox}

\subsection{Generalisation}
To test whether EvoML generalises beyond nnU-Net's medical grounding domain, we evaluate it on three general ML tasks: natural image classification (CIFAR-10~\cite{krizhevsky2014cifar}), natural image segmentation (VOC2012~\cite{pascal-voc-2012}), and text classification (AG News~\cite{AGscorpu44:online}), against all five baselines. These general tasks are comparatively simple and follow well-established pipelines, so all methods achieve near-perfect automation, with EAC close to $1.0$ throughout; Table~\ref{tab:rq4} therefore reports FMQ, the dimension on which they differ.

\begin{table}[!htbp]
\centering
\caption{Generalisation to three general ML tasks: FMQ (mean $\pm$ std).}
\label{tab:rq4}
\setlength{\tabcolsep}{5pt}
\begin{tabular}{@{}l ccc@{}}
\toprule
Method & CIFAR-10 & VOC2012 & AG News \\
\midrule
M3Builder    & 0.9216\,{\tiny$\pm$0.0049} & 0.6097\,{\tiny$\pm$0.0123} & 0.9113\,{\tiny$\pm$0.0044} \\
AutoML-Agent & 0.9419\,{\tiny$\pm$0.0027} & 0.7328\,{\tiny$\pm$0.0084} & 0.9103\,{\tiny$\pm$0.0056} \\
MLZero       & 0.9466\,{\tiny$\pm$0.0030} & 0.7117\,{\tiny$\pm$0.0068} & 0.9441\,{\tiny$\pm$0.0033} \\
Codex        & 0.9424\,{\tiny$\pm$0.0042} & 0.7036\,{\tiny$\pm$0.0088} & 0.9207\,{\tiny$\pm$0.0061} \\
Claude Code  & 0.9391\,{\tiny$\pm$0.0032} & 0.7086\,{\tiny$\pm$0.0065} & 0.9457\,{\tiny$\pm$0.0039} \\
\midrule
EvoML        & \textbf{0.9674}\,{\tiny$\pm$0.0026} & \textbf{0.7401}\,{\tiny$\pm$0.0058} & \textbf{0.9548}\,{\tiny$\pm$0.0030} \\
\bottomrule
\end{tabular}
% %\vspace{-5pt}
\end{table}
EvoML achieves the best FMQ across all three tasks, confirming that its framework generalises beyond medical imaging. 
The expert-grounded skills are not rigidly tied to the medical setting: the Executor adapts them to each new domain and post-run evolution refines them further, so the grounded procedures continue to benefit general tasks. 
The margins are narrower here than on the medical benchmarks largely because these tasks are simpler and more standardised, leaving less room to distinguish methods, yet EvoML's combination of adapted expert skills, validation-gated execution, and evolution still yields the strongest results on every task.
%\vspace{-5pt}
\begin{tcolorbox}[colback=gray!5,colframe=gray!50,boxrule=0.5pt,arc=10pt,left=2pt,right=2pt,top=2pt,bottom=2pt]
\textbf{Answer to RQ4:} EvoML extends beyond the targeted benchmarks to general ML tasks, achieving the best FMQ on all three with full automation. Its expert-grounded skills adapt to each task and, with validation-gated execution and evolution, continue to outperform every baseline.
\end{tcolorbox}

% -------------------------------------------------------------
\section{Discussion}
\label{sec:discussion}
% \subsection{Summary of findings}
\subsection{Architectural Comparison with Baselines}
\begin{table}[!htbp]
  \centering
  \caption{Architectural comparison of EvoML against agentic AutoML systems and command-line coding tools.}
  %\vspace{-5pt}
  \label{tab:design_comparison}
  \scriptsize
  \setlength{\tabcolsep}{3pt}
  \begin{tabular}{l c c c c}
    \toprule
    Method & \makecell{Manual-Artifact-\\Free Input} & \makecell{Expert\\Grounding} & \makecell{Stage-Gated\\Validation} & \makecell{Cross-run\\Evolution} \\
    \midrule
    M$^3$Builder & \makecell{\xmark\\\tiny(DataCard required)} & \makecell{\cmark\\\tiny(Fixed templates)} & \makecell{\xmark\\\tiny(Shallow output check)} & \xmark \\
    AutoML-Agent & \makecell{\xmark\\\tiny(Skeleton required)} & \xmark & \makecell{\xmark\\\tiny(Validation on\\\tiny pseudo execution)} & \xmark \\
    MLZero & \cmark & \makecell{\xmark\\\tiny(Document retrieval)} & \makecell{\xmark\\\tiny(Execution only)} & \xmark \\
    Codex & \cmark & \xmark & \makecell{\xmark\\\tiny(Execution only)} & \xmark \\
    Claude Code & \cmark & \xmark & \makecell{\xmark\\\tiny(Execution only)} & \xmark \\
    \midrule
    \textbf{EvoML} & \textbf{\cmark} & \cmark & \textbf{\cmark} & \textbf{\cmark} \\
    \bottomrule
  \end{tabular}
  % %\vspace{-5pt}
\end{table}
Table~\ref{tab:design_comparison} contrasts EvoML with the baselines on the architectural dimension that distinguishes it.
M3Builder and AutoML-Agent require a user-supplied skeleton or templates, whereas MLZero and the command-line tools run from minimal input but lack expert grounding. 
On validation, the baselines check only execution success (MLZero, command-line tools), file and shape structure (M3Builder), or conformance to user requirements assessed over pseudo-execution rather than real runs (AutoML-Agent); none verify task-level correctness, so a runnable but task-inappropriate pipeline passes unflagged. 
Furthermore, no baselines retain experience once a run completes. 
EvoML alone combines minimal-input operation, adaptive expert grounding, task-aware validation, and cross-run evolution, the combination that yields its automation, quality, and consistency.

\subsection{Cost Against Human Expert Development} 

\begin{table}[!htbp]
\centering
\caption{Cost-efficiency comparison against expert leaderboard solutions.}
%\vspace{-5pt}
\label{tab:cost}
\setlength{\tabcolsep}{6pt}
\renewcommand{\arraystretch}{0.9}
\begin{tabular}{lcc}
\toprule
Dataset & Human Leaderboard & EvoML \\
\midrule
\href{https://flare22.grand-challenge.org/evaluation/challenge/leaderboard/}{FLARE22} & 0.910 & \textbf{0.920} \\
\href{https://www.creatis.insa-lyon.fr/Challenge/acdc/results.html}{ACDC} & 0.936 & 0.919 \\
\href{https://www.synapse.org/Synapse:syn3193805/wiki/217785}{BTCV} & 0.921 & 0.837 \\
\href{https://flare.grand-challenge.org/evaluation/challenge/leaderboard/}{FLARE21} & 0.909 & \textbf{0.935} \\
\href{https://promise12.grand-challenge.org/evaluation/challenge/leaderboard/}{Promise12} & 0.919 & \textbf{0.920} \\
\href{https://medmnist.com/}{OrganMNIST3D} & 0.996 & 0.992 \\
\href{https://web.archive.org/web/20230324102125mp_/https://fastmri.org/leaderboards}{FastMRI} & 0.963 & 0.950 \\
\midrule
Time & 8--40 h & 20--45 min \\
Cost & \$800--\$6{,}000 & \$10--\$35 \\
\bottomrule
\end{tabular}
% %\vspace{-10pt}
\end{table}

We compare EvoML against expert-engineered leaderboard solutions for the evaluated datasets. As shown in Table~\ref{tab:cost}, EvoML surpasses the expert reference on FLARE21, FLARE22, and Promise12, remains close on OrganMNIST3D and FastMRI, but performs lower on ACDC and BTCV. 
Overall, EvoML achieves competitive model quality while substantially reducing synthesis effort and cost.
The comparison underscores EvoML's cost-efficiency: developing a leaderboard-level solution typically requires on the order of 8 to 40 hours of expert effort depending on task difficulty, excluding unattended model training, corresponding to roughly \$800 to \$6{,}000 in labour cost, whereas EvoML synthesises a pipeline in about 20--45 minutes at a generation cost of \$10 to \$35. It thus trades expensive, scarce expert time for cheap, fully automated synthesis while remaining competitive in quality.

\subsection{Threats to Validity}
\textbf{1. Internal validity:} EvoML depends on the underlying foundation model. To assess this effect, we rerun the full framework with Claude Opus 4.7 on BTCV, FLARE22, and FastMRI. The resulting FMQ differs from GPT-5.5 by less than 0.008, suggesting that the main findings are not specific to one backbone. Since Opus 4.7 costs roughly 1.7× as much per pipeline, we use GPT-5.5 as the default backbone for comparable quality at a lower synthesis cost.
\textbf{2. Construct validity:} EvoML incurs longer model-training time because its nnU-Net grounding includes thorough preprocessing, augmentation, and five-fold cross-validation. On a typical supervised segmentation task, the baselines train in roughly 10 hours, EvoML in 60, and nnU-Net in 300. We consider this additional computation a trade-off for the stronger automation and model quality enabled by the grounded workflow.
\textbf{3. External validity:} We focus primarily on medical imaging because its pipelines rely on domain-specific processing and modelling choices, where mistakes can remain executable yet still degrade final model quality. We additionally evaluate three general ML tasks as initial evidence beyond this setting. For broader SE tasks, the same principles remain applicable: workflows can be decomposed into stages, recurring procedures can be encoded as skills, intermediate artifacts can be validated before downstream execution, and successful recoveries can be retained across tasks. Extending EvoML therefore mainly requires adapting the skill content and validation criteria.
% -------------------------------------------------------------

\section{Conclusion}
\label{sec:conclusion}
We presented EvoML, an agentic framework for end-to-end ML pipeline synthesis built on expert-grounded skills, validation-gated execution, and post-run skill evolution. By advancing each phase only when its outputs satisfy structural, execution, and semantic checks, EvoML limits error propagation, while post-run evolution consolidates successful recoveries into persistent skill updates that improve execution across subsequent tasks. Across eight medical-imaging datasets and three general ML tasks, EvoML achieved full automation and the strongest, most consistent model quality among the compared systems, while substantially reducing synthesis effort relative to expert development. Our additional evolution analysis further shows improved first-pass skill success, fewer validation-triggered retries, and cross-task reuse of retained adaptations. Future work will extend EvoML to broader ML and software-engineering tasks and investigate finer-grained validation throughout the pipeline to further improve model quality and reliability.

\section{Data Availability}
The artifacts that support the results discussed in this paper are available at:
\url{https://anonymous.4open.science/r/EvoML-624B}.

\bibliographystyle{IEEEtran}
\bibliography{bibi}

\end{document}